%% file: iclr2027_conference.tex
\documentclass{article} 
\usepackage{iclr2027_conference,times}

\input{math_commands.tex}

\usepackage{hyperref}
\usepackage{url}
\usepackage{xcolor}
\usepackage{xspace}
\usepackage{graphicx}
\usepackage{booktabs}
\usepackage{amsmath}
\usepackage{multirow}
\usepackage{placeins}

\newcommand{\method}{\textsc{DeOrch}\xspace}

\title{Decoupled Multi-Agent Orchestration}

\author{Xinle Wu, Yao Lu \\
National University of Singapore\\
\texttt{\{wuxl,luyao\}@comp.nus.edu.sg} \\
}

\iclrfinalcopy 
\begin{document}

\maketitle
\lhead{Under review as a conference paper at ICLR 2027}

\begin{abstract}

Learned orchestration can automatically construct effective language-model multi-agent systems, but existing approaches couple planning to fixed worker pools and train decomposition and collaboration from the same terminal outcome, limiting transfer and obscuring credit assignment. We introduce \method, which separates worker-agnostic planning from concrete worker selection. Its two-stage planner first decomposes the task without worker information, then chooses collaboration operations using compact, worker-identity-free matchability feedback from the pool, enabling conditional credit assignment to decomposition and collaboration decisions. A lightweight matcher estimates worker suitability from behavior on a fixed probe set and adapts online with a contextual bandit, allowing new workers to be incorporated without retraining the planner or matcher. Across diverse in- and out-of-distribution tasks, \method outperforms prior automatic MAS orchestration methods with fewer worker calls than competing learned orchestrators, remains effective when transferred to an entirely unseen worker pool without retraining, and shows consistent gains from both components.

\end{abstract}

\section{Introduction}
\label{sec:intro}

A multi-agent system (MAS) tackles a task with several language-model agents rather than one alone, letting them divide the work, verify one another, and combine their complementary strengths. Studies have shown that on some complex tasks a MAS outperforms each of its constituent agents~\citep{wang2025mixture,kim2025towards,nielsen2026learning}. An expert-designed collaboration structure, specifying how the task is divided and how the agents work on it, often performs well; hand-designing one for every task, however, is impractical. Recent work therefore learns to build it automatically~\citep{zhang2024g,dang2025multiagent,gao2025flowreasoner,cui2026uno}.

A recent line of work trains an orchestrator that, given a task, decides how to decompose it into subtasks and how each subtask is handled by worker agents in a fixed pool~\citep{gao2025flowreasoner,chi2025era,cui2026uno}. For instance, Conductor~\citep{nielsen2026learning} learns, over a pool of powerful frozen workers, to decompose a task, assign each subtask to a worker, and specify how intermediate outputs are shared, training the orchestrator end-to-end on final task correctness. Relative to Conductor, MAS-Orchestra~\citep{ke2026mas} does not model worker selection; it instead uses a fixed underlying worker and learns only to decompose the task and choose an operation for each subtask (e.g., a single worker, an ensemble, or a debate). Trained on outcome reward alone, these orchestrators learn effective coordination: by directing a fixed set of strong workers, a small orchestrator makes them collectively surpass any individual worker, and even rival much larger models, at a fraction of the cost.

Despite these successes, current automatic-MAS methods suffer from two limitations that reduce their effectiveness and practicality in realistic settings.
\textbf{(1) The trained orchestrator struggles to exploit unseen workers.} Existing methods either ignore concrete worker selection or learn it directly into the orchestrator's policy~\citep{ke2026mas,nielsen2026learning,gao2025flowreasoner}. The former cannot exploit differences in workers' capabilities, while the latter cannot generalize its learned worker-selection strategy to unseen workers. This limitation is increasingly relevant in practice as new and stronger agents continually emerge: as the worker pool evolves, the orchestrator is either blind to which workers are best suited to each subtask or must be adapted to exploit newly available workers.
\textbf{(2) A single outcome reward confounds credit assignment between decomposition and operation decisions.} The orchestrator makes two coupled decisions: how to decompose the task and which operation to run on each subtask~\citep{ke2026mas,cui2026uno}. A failure can stem from either one: the same decomposition may succeed when a subtask is handled by an ensemble but fail when that subtask is left to a single worker, yet a single outcome reward cannot isolate which decision is responsible, leaving both with entangled learning signals. Moreover, once the subtasks are determined, the appropriate operation for each one depends on how well the available workers match that specific subtask: several moderately matched workers may favor an ensemble, whereas one clearly dominant match may favor a single worker. Without such subtask-specific information about the worker pool, the orchestrator cannot reliably distinguish these cases.

In this paper, we introduce \method, an automatic MAS orchestration framework that addresses both limitations by decoupling planning from concrete worker selection. Specifically, \method consists of two parts.
\textbf{(1) A worker-agnostic, two-stage planner with matchability feedback.} The planner makes two decisions: how to decompose the task into subtasks and which collaboration operation to run on each. Rather than making both decisions at once, it proceeds in two stages. It first decomposes the task without observing the workers; the matcher then summarizes, for each subtask, how well the worker pool can support it through a compact, worker-identity-free \emph{matchability} profile. Conditioned on this feedback, the planner chooses a collaboration operation for each subtask while keeping the decomposition fixed. This factorization enables separate credit assignment to decomposition and operation decisions during training, rather than training both from the same confounded outcome signal. It also localizes worker dependence: task decomposition is performed without worker information, while collaboration decisions adapt to the current pool through matchability feedback. Concrete worker selection remains entirely in the matcher, allowing the same planner to be applied to changed worker pools without retraining.
\textbf{(2) A lightweight matcher combining probe-based profiling with online adaptation.} For each \emph{(subtask, operation)} pair, the matcher selects suitable workers using both probe-based prior estimates of their capabilities and execution feedback collected during deployment. It first evaluates each worker on a fixed probe set, producing a behavioral profile from which its suitability for a new subtask is estimated based on semantically related probes. This provides a transferable cold-start estimate for new workers. A contextual bandit then refines these estimates from observed execution outcomes during deployment. Because worker-specific knowledge is confined to the matcher, a new or upgraded worker can be incorporated simply by evaluating it on the probe set, without retraining the planner or matcher.

We evaluate \method across diverse in- and out-of-distribution tasks spanning math, code, multi-hop QA, tool use, and planning. \method consistently outperforms existing automatic MAS orchestration baselines while using fewer worker calls, and remains effective when transferred to an entirely unseen worker pool without retraining. 
Ablations further validate both components: two-stage planning, matchability feedback, and conditional credit assignment improve orchestration, while probe-based matching, complementarity-aware worker selection, and online adaptation improve worker selection.

%

%

%
\section{Related Work}
\label{sec:related}

\paragraph{Automatic MAS Orchestration}
Automatic MAS design replaces hand-crafted collaboration with systems that automate decisions such as task decomposition, agent roles and prompts, and the flow of information among agents \citep{wang2026mas,li2026agentswift,li2026ofa,li2025agent,shang2025agentsquare}. One line uses off-the-shelf LLMs to design or search over MAS structures, without training a dedicated orchestration model \citep{zhang2025aflow,fourney2024magentic,li2026experience,lin2026unified,liu2026mas,yun2026graph,zhang2025metaagent,wang2026metagen,niu2025flow,ke2025mas}. For example, \textsc{MetaAgent}~\citep{zhang2025metaagent} asks an LLM to instantiate a set of agents and construct a finite-state machine that routes among them, while \textsc{MASS}~\citep{zhou2026multi} optimizes both collaboration topologies and the prompts within them through a three-stage search pipeline. A second line uses non-LM controllers or optimization mechanisms to choose the MAS structure \citep{jiang2026agent,li2026multi,xu2026evomas,li2026ofa,li2026assemble,wu2026card,wang2026neuralfsm,leong2025amas}. For example, \textsc{G-Designer}~\citep{zhang2024g} uses a GNN-based generator to decide which agents should communicate for a given task, while \textsc{MaAS} \citep{zhang2025multi} learns a supernet containing many candidate agentic operators and activates only the parts needed for each query. These methods primarily optimize structural decisions rather than generating free-form task decompositions and subtask instructions in language. A third and more closely related line trains a language model specifically for orchestration \citep{dang2025multiagent,wang2026agentconductor,cui2026uno,chi2025era,ruan2026aorchestra,huangmas}. For example, \textsc{Conductor}~\citep{nielsen2026learning} uses reinforcement learning to train an LM that jointly writes the subtasks and assigns each subtask to a worker. \textsc{MAS-Orchestra}~\citep{ke2026mas} emits the entire orchestration in one decision step, instantiating for each subtask one of a fixed set of sub-agent types such as self-consistency, debate, or self-refine. \textsc{FlowReasoner}~\citep{gao2025flowreasoner} generates for each query a Python program defining the flow of a multi-agent system, with its subtasks handled by the same worker model under different roles. 
%
Unlike these approaches, which either omit concrete worker selection or couple it to planning, \method keeps the planner worker-agnostic and delegates worker selection to a separate matcher.

\paragraph{LLM Routing and Model Selection}

LLM routing selects an appropriate model for each query from a candidate pool~\citep{lu2024routing,tsiourvas2025causal,mei2025omnirouter}. Many existing routers are trained offline. One line trains a parametric router to predict which model best matches the input query~\citep{ong2025routellm,chen2024routerdc,song2025irt,somerstep2025carrot}. For example, \textsc{P2L}~\citep{frick2025prompt} trains a language-model router whose output scores correspond to the candidate models, tying the router itself to the model pool. Another line avoids training such a router and instead estimates a model's performance on a new query from its behavior on a reference set~\citep{shadid2025ori,jitkrittum2026universal,li2025rethinking}. \textsc{Avengers-Pro}~\citep{zhang2025beyond}, for example, maps a new query to semantically similar query clusters and routes according to the models' observed performance on them; a newly added model only needs to be evaluated on the reference set. A separate line studies online routing, treating model selection as a contextual-bandit problem and updating the router from deployment feedback, for example with LinUCB or Thompson sampling~\citep{nguyen2025laser,li2025llm}. 

\paragraph{Credit Assignment in RL for LLMs}
Credit assignment arises whenever a single outcome reflects multiple decisions~\citep{pignatelli2023survey}. In LLM reasoning, an incorrect final answer gives little indication of which intermediate reasoning steps were useful or harmful, motivating process reward models~\citep{lightman2024let}. In cooperative multi-agent learning, a shared team reward similarly obscures the contribution of individual agents, motivating value decomposition and counterfactual agent-level credit~\citep{liang2024asynchronous,zhao2025multi}. The same ambiguity appears in learned MAS orchestration, where a terminal outcome can entangle planning with worker selection and, within the planner, can further entangle task decomposition with operation decisions; recent work such as \textsc{LEMON}~\citep{chen2026lemon} localizes credit by counterfactually editing individual orchestration fields. 
%
To our knowledge, \method is the first orchestration framework to use two-stage planning for RL credit assignment, separating decomposition from collaboration so that they receive distinct learning signals.

\section{Method}
\label{sec:Method}

Figure 1 gives an overview of \method, which separates MAS orchestration into a worker-agnostic task planner and a separate matcher responsible for worker selection. Given a task, the language-model planner first decomposes it into a DAG of subtasks, whose edges encode dependencies. The matcher then summarizes how well the current worker pool can support each subtask through worker-identity-free matchability features, which the planner uses in a second pass to assign collaboration operations while keeping the decomposition fixed. Finally, the matcher binds each planned operation to suitable workers using its probe-based prior and online adaptation, and the resulting MAS is executed; execution outcomes are used to train the planner and update the matcher.


\begin{figure*}[t]
  \centering
    \includegraphics[width=0.90\linewidth]{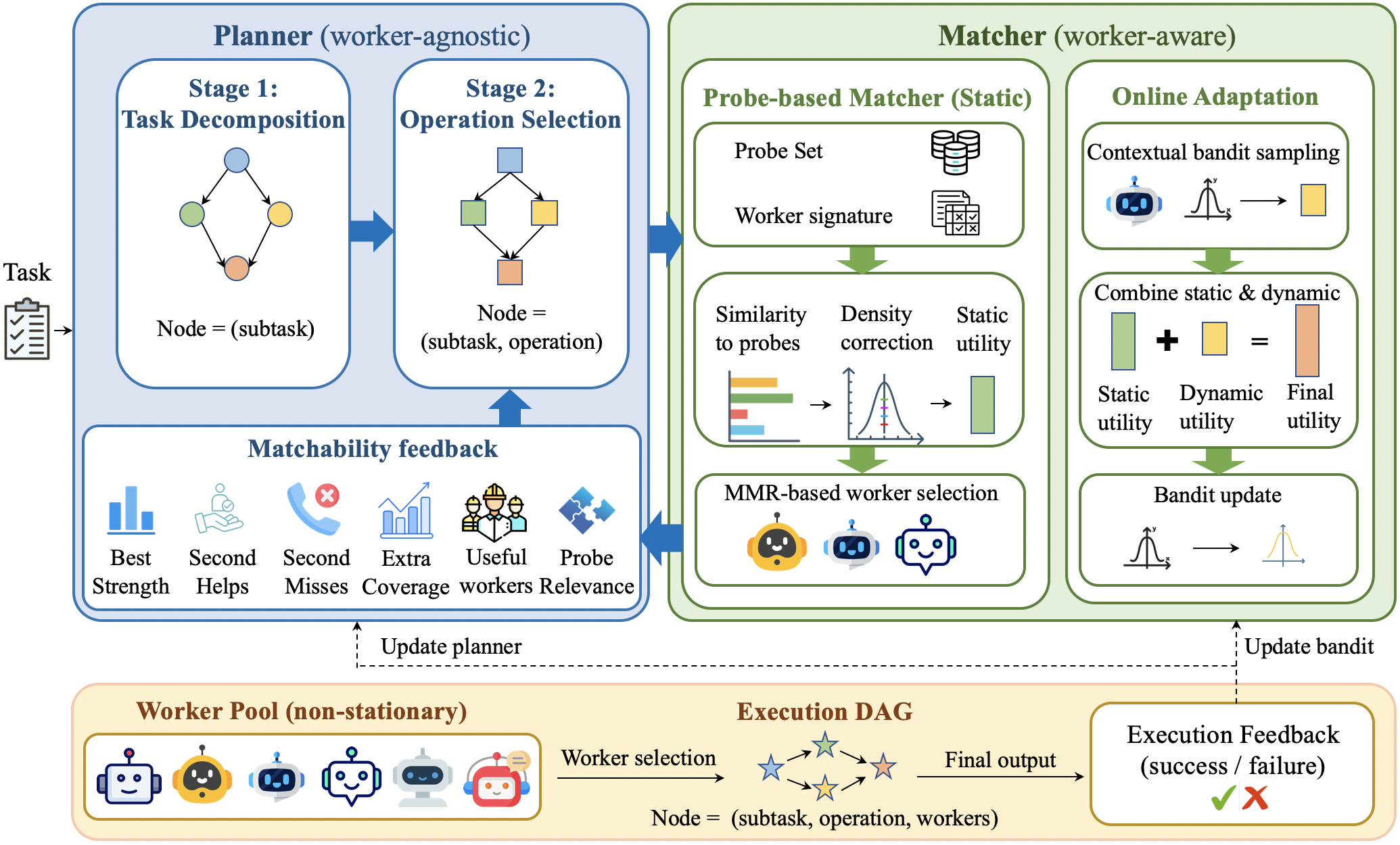}
  \caption{Overview of \method.
  }
  \label{fig:overview}
\end{figure*}

\subsection{Problem Formulation}
\label{sec:formulation}

Given a query $x$ and a pool of worker agents $\mathcal{W}=\{w_1,\ldots,w_M\}$, automatic MAS orchestration constructs a multi-agent workflow for solving $x$. We represent an orchestration by
\[
\mathcal{A}
=
\left(
D=(S,E),
\{(o_s,n_s)\}_{s\in S},
\{B_s\}_{s\in S}
\right),
\]
where $D$ is a task-decomposition DAG whose nodes $S$ are subtasks decomposed from $x$ and whose directed edges $E$ specify their dependencies. For each subtask $s$, $o_s\in\mathcal{O}$ denotes a collaboration operation from $\{\textsc{Single}, \textsc{Ensemble}, \textsc{Debate}, \textsc{Select}, \textsc{Review}\}$, and $n_s$ its required number of workers, while $B_s\subseteq\mathcal{W}$ denotes the concrete worker team assigned to execute it, with $|B_s|=n_s$. Executing $\mathcal{A}$ produces a final output $y$, whose quality is measured by an evaluator $Q(x,y)$. 
The orchestration problem therefore involves deciding how to decompose the task, how each subtask should be collaboratively executed, and which available workers should carry out that execution. 


\subsection{Worker-Agnostic Two-Stage Planner}
\label{sec:planner}


Existing LM orchestrators often couple task planning with worker selection in the same policy~\citep{nielsen2026learning,cui2026uno}, tying the orchestrator to a particular worker pool and requiring retraining when the pool changes. Our planner is instead worker-agnostic: it decides how to decompose the task and how each subtask should be collaboratively executed, while all concrete worker selection is delegated to the matcher. Workers can therefore be added to or removed from the pool without retraining the planner.
The planner operates in two stages, separating task decomposition from collaboration planning for two reasons. First, a shared terminal reward entangles the two decisions: failure does not reveal whether the decomposition or collaboration strategy was responsible. Separating the two decisions allows collaboration choices to be sampled and compared while holding the decomposition fixed, yielding more targeted credit for collaboration decisions during RL training. Second, collaboration should depend on how well the current worker pool supports each subtask. For example, a subtask that is easily handled by the available workers may require only a single worker, whereas a poorly matched subtask may benefit from ensemble execution. The second stage therefore conditions operation selection on worker-identity-free matchability feedback, adapting collaboration to the pool without exposing worker identities.

\paragraph{Stage I: task decomposition.}
Given an input query $x$, the planner generates a task-decomposition DAG
$D=(S,E)\sim\pi_{\theta}^{\mathrm{dec}}(\cdot\mid x)$,
where $S=\{s_1,\ldots,s_L\}$ contains natural-language subtasks and $E$ specifies their dependencies.
At this stage, the planner decomposes the task from the query alone, determining what should be solved and in what dependency order, while leaving collaboration planning and worker selection to subsequent stages.

\paragraph{Stage II: collaboration planning.}
After generating the decomposition, the matcher evaluates how well the worker pool supports each subtask and returns a matchability feature $m_s = m(s;\mathcal{W})$, a fixed-dimensional, worker-identity-free summary of pool support (Section~\ref{sec:matcher}). Conditioned on the query, fixed decomposition, and matchability features, the planner performs a second generation:
\[
    \{(o_s,n_s)\}_{s\in S}
    \sim
    \pi_{\theta}^{\mathrm{op}}
    \left(
        \cdot
        \mid
        x,D,\{m_s\}_{s\in S}
    \right),
\]
where $o_s\in\mathcal{O}$ specifies the collaboration operation used for subtask $s$ and $n_s$ denotes the required team size. Stage II keeps the decomposition fixed and augments each subtask with $(o_s,n_s)$, adapting collaboration to the current pool through matchability feedback while leaving concrete worker selection to the matcher.

\paragraph{Collaboration operations.}
We consider a compact operation set
$\mathcal{O} = \{\textsc{Single}, \textsc{Ensemble}, \textsc{Debate}, \textsc{Select}, \textsc{Review}\}$.
Each operation determines how the assigned worker or worker team interacts when solving $s$. Specifically, \textsc{Single} uses one worker; \textsc{Ensemble} aggregates responses from multiple workers; \textsc{Debate} lets multiple workers exchange and refine their responses; \textsc{Select} uses an adjudicator to choose among candidate responses; and \textsc{Review} uses reviewers to critique and refine an initial response. By composing these operations over the subtask DAG, the planner can instantiate common MAS structures.


\subsection{Probe-Based Matcher with Online Adaptation}
\label{sec:matcher}


The matcher combines a static probe-based matcher with an online contextual bandit. The static matcher estimates worker suitability for a subtask from performance on semantically related probes, while the online bandit refines these estimates from deployment feedback. The static matcher is used for planner training and deployment, while online adaptation is used only at deployment.

\paragraph{Probe-based matching with density correction.}

We construct a fixed probe set of $N$ diverse tasks. Each worker $w$ is evaluated once on the same probes, producing a behavioral signature $\mathbf{o}_w=(o_{w,1},\ldots,o_{w,N})\in\{0,1\}^{N}$, where $o_{w,j}=1$ if worker $w$ succeeds on probe $j$ and $0$ otherwise.
Given a subtask $s$, we estimate worker suitability from its outcomes on semantically related probes. Let $e_s$ and $e_j$ denote the embeddings of $s$ and probe $j$. We define the subtask--probe relevance as
$\kappa_j(s)=\exp((\cos(e_s,e_j)-1)/h_{\mathrm{rel}})$,
where smaller $h_{\mathrm{rel}}$ focuses on more similar probes. Directly normalizing these weights can overemphasize semantic regions containing many near-duplicate probes. We therefore estimate the local probe density
\[
    q_j
    =
    \sum_{\ell=1}^{N}
    \exp\left(
        \frac{\cos(e_j,e_\ell)-1}{h_{\mathrm{den}}}
    \right),
    \qquad
    \bar q_j
    =
    \frac{q_j}{\operatorname{median}_{\ell} q_\ell},
\]
where $h_{\mathrm{den}}$ controls the neighborhood scale. The final probe weights and worker suitability are
\[
    \alpha_j(s)=
    \frac{\kappa_j(s)(\bar q_j+\epsilon)^{-\nu}}
    {\sum_{\ell=1}^{N}\kappa_\ell(s)(\bar q_\ell+\epsilon)^{-\nu}},
    \qquad
    \hat u_w(s)=\sum_{j=1}^{N}\alpha_j(s)o_{w,j}.
\]
The density correction reduces the influence of locally redundant probes while preserving genuinely distinct coverage. Since $\alpha_j(s)$ depends only on the subtask and the fixed probe set, a new worker can be scored directly from its probe signature without retraining the matcher.

\paragraph{Matchability feedback.}
The second-stage planner should know how well the current worker pool can support a subtask without observing worker identities. We therefore summarize the pool with a fixed six-dimensional feature
$m(s)=[B(s),U(s),L(s),G(s),D(s),C(s)]$. The six coordinates capture complementary aspects of pool capability: $B$ measures the strength of the best-matched single worker; $U$ and $L$ describe the benefit and risk of adding a complementary second worker; $G$ and $D$ summarize the amount and breadth of complementary capability available across additional workers; and $C$ measures how well the probe set supports the estimate for the current subtask. Importantly, these statistics describe the pool without encoding worker identities or its size. Hence the planner receives the same fixed interface as workers are added or removed, while the matcher recomputes the summary from the current pool.

Let $w_1=\arg\max_w \hat u_w(s)$ be the best-matched worker, so that $B(s)=\hat u_{w_1}(s)$. Starting from $w_1$, we greedily add workers according to their marginal increase in weighted probe coverage, up to a maximum team size $K$. Let $g_k$ denote the marginal coverage gain from the $k$-th addition. The first added worker, denoted $w_2$,
maximally covers the probe failures of $w_1$, giving
\[
    U(s)=\sum_j\alpha_j(s)(1-o_{w_1,j})o_{w_2,j},
    \qquad
    L(s)=\sum_j\alpha_j(s)o_{w_1,j}(1-o_{w_2,j}),
\]
where $U$ measures rescued failure mass and $L$ measures converse spoilage mass. In particular, $g_1=U(s)$. The resulting gains $g_1,\ldots,g_{K-1}$ are summarized as
\[
    G(s)=\sum_k g_k,
    \qquad
    D(s)=\frac{G(s)^2}{\sum_k g_k^2},
    \qquad
    C(s)=\sum_j\alpha_j(s)\cos(e_s,e_j).
\]
Here, $G$ measures the total complementary coverage available beyond the best single worker, $D$ reflects how broadly that coverage is distributed across additional workers, and $C$ measures the semantic support provided by the probes used for the estimate.

\paragraph{Operation-specific worker selection.}
Once the planner specifies $(s,o_s,n_s)$, the matcher selects workers using subtask suitability and behavioral complementarity. For \textsc{Single}, we select the worker with the highest suitability score $\hat u_w(s)$. For multi-worker operations, we use maximal marginal relevance (MMR)~\citep{carbonell1998use} for team construction. We initialize $B$ with the highest-suitability worker and then greedily add workers until $|B|=n_s$. 
%
We define the directional complementarity of workers $w$ and $v$ and select each additional worker according to
\[
\delta_{w\leftarrow v}(s)
=
\sum_{j=1}^N \alpha_j(s)(1-o_{v,j})o_{w,j},
\qquad
w^\star
=
\arg\max_{w\notin B}
\left[
\lambda_o\hat u_w(s)
-
(1-\lambda_o)\max_{v\in B}(1-\delta_{w\leftarrow v}(s))
\right].
\]
This criterion balances competence with failure-region complementarity to the
selected team $B$. We use it for \textsc{Ensemble} and \textsc{Debate}.
Operations with asymmetric roles use specialized rules: \textsc{Select} first assigns the highest-suitability worker as the adjudicator and uses MMR to choose the remaining proposers, whereas \textsc{Review} assigns the highest-suitability worker as the producer and greedily selects reviewers that best cover its failure region.

\paragraph{Online adaptation.}
%
The probe-based matcher provides a cold-start estimate from worker behavior, while deployment experience provides evidence of worker utility on the target distribution. We therefore augment the static matcher with contextual Thompson sampling. Each worker $w$ has a probe-derived capability representation $g(w)$, and we maintain a Gaussian posterior over a shared parameter $\eta$ modeling context-dependent task--worker interactions. Before each MAS execution, we sample $\tilde{\eta}$ from the posterior and combine the probe-based estimate with the interaction between the subtask representation $z(s)$ and worker representation $g(w)$ to form the deployment-adapted suitability $\tilde u_w(s)=\hat u_w(s)+\tilde{\eta}^\top(z(s)\odot g(w))$.
At deployment, $\tilde u_w(s)$ replaces $\hat u_w(s)$ in the operation-specific worker-selection rules, while the probe-derived complementarity statistics remain unchanged. After the MAS is executed, its terminal utility is treated as noisy bandit feedback for the selected worker assignment and used to update the shared posterior. The bandit does not directly attribute a single outcome to individual workers; rather, repeated feedback across tasks and worker assignments allows deployment-specific task--worker interactions to be learned over time.

\subsection{Planner Training with Conditional Credit Assignment} \label{sec:training} 

We post-train the planner from terminal execution outcomes using nested conditional rollouts. For each query, we sample multiple decompositions and, for each fixed decomposition, multiple collaboration plans. Comparing collaboration plans under the same decomposition provides credit for Stage II, while averaging their outcomes estimates how good that decomposition is for Stage I. This yields separate learning signals for the two planning stages from the same terminal outcomes.

\paragraph{Conditional credit assignment.}

For each training query $x$, we sample $J$ decompositions and, for each fixed decomposition, $k$ collaboration plans:
\[
\begin{aligned}
D_j
&\sim
\pi_\theta^{\mathrm{dec}}(\cdot\mid x),
&& j=1,\ldots,J,\\
O_{j,l}
&\sim
\pi_\theta^{\mathrm{op}}
\bigl(\cdot\mid x,D_j,\{m_s\}_{s\in S_j}\bigr),
&& l=1,\ldots,k.
\end{aligned}
\]
Each resulting plan is assigned concrete workers by the matcher and executed, yielding terminal quality $Q_{j,l}=Q(x,y_{j,l})$.
For Stage II, all $k$ rollouts under the same $D_j$ share the same upstream decomposition, so their relative outcomes provide credit for the collaboration decisions. We first form a cost-adjusted reward and then compute its relative advantage within the group:
\[
    R^{\mathrm{op}}_{j,l}
    =
    Q_{j,l}\exp(-\beta c_{j,l}),
    \qquad
    A^{\mathrm{op}}_{j,l}
    =
    R^{\mathrm{op}}_{j,l}
    -
    \frac{1}{k}
    \sum_{l'=1}^{k}
    R^{\mathrm{op}}_{j,l'},
\]
where $c_{j,l}$ is the number of worker calls incurred by rollout $(j,l)$ and $\beta$ controls the cost penalty. For Stage I, we first average over downstream collaboration choices and then compare the resulting decomposition-level scores:
\[
    \bar Q_j
    =
    \frac{1}{k}
    \sum_{l=1}^{k}
    Q_{j,l},
    \qquad
    A^{\mathrm{dec}}_j
    =
    \bar Q_j
    -
    \frac{1}{J}
    \sum_{j'=1}^{J}
    \bar Q_{j'}.
\]
We use $A^{\mathrm{op}}$ and $A^{\mathrm{dec}}$ to optimize the corresponding Stage-II and Stage-I generations, respectively.

\paragraph{Cost-aware collaboration planning.}
Existing learned orchestrators such as \textsc{Conductor} and \textsc{MAS-Orchestra} primarily optimize task quality without explicitly accounting for collaboration cost. Since more intensive collaboration typically incurs more worker calls and can also improve performance, a quality-only objective may systematically favor heavier operations. This cost--quality tradeoff is directly controlled by Stage II. In contrast, Stage I determines the task decomposition, and decomposition granularity by itself does not indicate solution quality: finer or coarser decompositions can each be preferable depending on the task. Our two-stage formulation therefore allows us to train Stage I with task quality alone while applying the cost-aware objective only to Stage II. The multiplicative reward further ensures that lower cost is valuable only when accompanied by task success: a failed rollout does not receive a higher reward merely because it uses fewer worker calls.

\section{Experiments}
\label{sec:experiments}

\subsection{Experimental Setup}
\label{sec:setup}

\paragraph{Models.}
We initialize the planner from \texttt{Qwen3-8B} and perform full-parameter RL post-training. Unless otherwise specified, all methods operate over the same heterogeneous pool of six worker models: \texttt{Qwen3-4B-Instruct-2507}, \texttt{gemma-4-12B-it}, \texttt{NVIDIA-Nemotron-Nano-9B-v2}, \texttt{Qwen3.5-9B}, \texttt{granite-4.1-8b}, and \texttt{Ministral-3-14B-Instruct-2512}.
We use \texttt{Qwen3-Embedding-8B} as the text encoder for semantic matching and retrieval.


\paragraph{Data and benchmarks.}
We use disjoint data for probe construction, planner post-training, and final evaluation. We evaluate in-distribution on held-out MuSiQue and TravelPlanner examples, and out-of-distribution on Omni-MATH-2, LiveCodeBench v6, BBEH, NaturalPlan, PlanBench-mystery, and GAIA, spanning math, code, multi-hop reasoning, tool use, and planning. Probe and training datasets, data sources, splits, and sample counts are provided in Appendix~\ref{app:dataset}.

\paragraph{Baselines.}
We compare against several complementary baselines. \textsc{Best Single Agent} uses \texttt{gemma-4-12B-it}, the best-performing worker in hindsight. As a fixed hand-designed MAS baseline, we include \textsc{Independent} coordination from prior work~\citep{kim2025towards}. \textsc{Frontier-Planner} uses Qwen3.8-27B as an off-the-shelf planner, with the rest of the system following \method. \textsc{Untrained Planner} uses the same planner initialization as \method but without orchestration-specific RL post-training. We further compare with learned automatic orchestrators, including \textsc{Conductor}~\citep{nielsen2026learning} and \textsc{MAS-Orchestra}~\citep{ke2026mas}, adapted to our worker setting. Additional baseline details are provided in Appendix~\ref{app:details}.

\subsection{Main Results}
\label{sec:main_results}

Table~\ref{tab:main} summarizes the main results. \method achieves the best overall performance across both in-distribution and out-of-distribution benchmarks. It substantially outperforms \textsc{Best Single Agent} and \textsc{Independent}, showing that its gains cannot be explained by relying on the strongest worker or by simply executing all workers in parallel. Both \textsc{Frontier-Planner} and \textsc{Untrained Planner} perform considerably worse, indicating that strong general-purpose planning or an untrained planner does not substitute for orchestration-specific post-training. Compared with learned automatic orchestrators, \method consistently outperforms \textsc{Conductor} and \textsc{MAS-Orchestra}, demonstrating the effectiveness and competitiveness of our approach. Moreover, \method uses fewer worker calls than both learned orchestrators, indicating that its performance gains do not come from making more worker calls.

\begin{table*}[htbp]
\centering
\caption{
Main results across in-distribution and out-of-distribution benchmarks.
}
\small
\setlength{\tabcolsep}{3.3pt}
\begin{tabular}{lcc|cccccc|cc}
\toprule
& \multicolumn{2}{c|}{In-Distribution}
& \multicolumn{6}{c|}{Out-of-Distribution}
& \multicolumn{2}{c}{Overall} \\
\cmidrule(lr){2-3}
\cmidrule(lr){4-9}
\cmidrule(lr){10-11}
Method
& MuSiQue & TP
& Omni & LCB & BBEH & NP & PB-M & GAIA
& Avg. & Calls $\downarrow$ \\
\midrule

\textsc{Best Single Agent}
& 72.8 & 13.6 & 42.5 & 69.9 & 38.4 & 51.1 & 19.4 & 6.1 & 39.2 & 1.00 \\

\textsc{Independent}
& 72.0 & 9.2 & 47.4 & 76.4 & 45.2 & 63.1 & 25.9 & 28.5 & 46.0 & 7.00 \\

\textsc{Frontier-Planner}
& 59.1 & 3.5 & 38.9 & 71.0 & 37.4 & 45.5 & 9.1 & 28.5 & 36.6 & 4.67 \\

\textsc{Untrained Planner}
& 60.2 & 3.3 & 38.1 & 70.4 & 31.8 & 50.4 & 25.2 & 26.7 & 38.3 & 4.92 \\

\midrule

\textsc{Conductor}
& 70.3 & 9.9 & 41.2 & 75.0 & 37.1 & 52.9 & 38.0 & 27.9 & 44.0 & 6.16 \\

\textsc{MAS-Orchestra}
& 73.4 & 15.1 & 49.0 & 76.9 & 45.0 & 61.8 & 41.6 & 29.7 & 49.1 & 5.63 \\

\midrule

\method
& 76.9 & 17.8 & 51.7 & 79.2 & 47.9 & 64.7 & 43.1 & 30.9 & 51.5 & 5.35 \\

\bottomrule
\end{tabular}
\label{tab:main}
\end{table*}

\subsection{Generalization to an Unseen Worker Pool}
\label{sec:pool_generalization}

%
Table~\ref{tab:pool_generalization} evaluates generalization to an entirely unseen four-worker pool (Appendix~\ref{app:details}), with no planner or matcher retraining and online adaptation disabled. \method remains effective, outperforming both \textsc{Best Single Agent} and \textsc{Independent} by clear margins. In contrast, \textsc{Conductor} degrades from 44.0 to 40.9: its advantage over \textsc{Best Single Agent} nearly disappears, its gap to \textsc{Independent} widens, and its gap to \method increases from 7.5 to 9.3 points. These results support the benefit of decoupling planning from concrete worker selection.

\begin{table*}[htbp]
\centering
\caption{
Performance under an unseen worker pool.
}
\small
\setlength{\tabcolsep}{3.3pt}
\begin{tabular}{lcc|cccccc|c}
\toprule
& \multicolumn{2}{c|}{In-Distribution}
& \multicolumn{6}{c|}{Out-of-Distribution}
& \multicolumn{1}{c}{Overall} \\
\cmidrule(lr){2-3}
\cmidrule(lr){4-9}
\cmidrule(lr){10-10}
Method
& MuSiQue & TP
& Omni & LCB & BBEH & NP & PB-M & GAIA
& Avg. \\
\midrule

\textsc{Best Single Agent}
& 69.0 & 6.2 & 45.9 & 75.9 & 47.7 & 40.3 & 29.0 & 12.1 & 40.8 \\

\textsc{Independent}
& 72.4 & 5.0 & 50.6 & 81.2 & 48.4 & 60.4 & 41.6 & 20.0 & 47.5 \\

\textsc{Frontier-Planner}
& 61.4 & 3.1 & 44.4 & 76.3 & 46.8 & 52.2 & 18.1 & 21.8 & 40.5 \\

\textsc{Conductor}
& 63.7 & 4.6 & 45.5 & 75.5 & 46.2 & 48.8 & 23.7 & 18.8 & 40.9 \\

\method
& 75.5 & 11.3 & 52.3 & 83.0 & 49.7 & 62.3 & 44.3 & 23.0 & 50.2 \\

\bottomrule
\end{tabular}
\label{tab:pool_generalization}
\end{table*}

\subsection{Analysis of the Planner}
\label{sec:credit_analysis}

%
%
We construct controlled variants to isolate the contributions of the planner design and training objective. \textsc{Joint Planning} generates decomposition and collaboration decisions in one pass under a shared terminal reward. \textsc{Two-Stage} separates the two decisions but uses neither matchability feedback nor conditional credit. \textsc{+ Matchability} provides the Stage-II planner with $m(s)$, and \textsc{+ Conditional Credit} further introduces our nested credit assignment. The full \method additionally applies the cost-aware Stage-II utility. We also evaluate \textsc{Decomposition + Single}, which preserves the learned decomposition but fixes every subtask to \textsc{Single}, measuring how much performance remains without Stage-II collaboration choices. All trained variants use the same planner backbone, worker pool, and rollout budget.
%
%
Table~\ref{tab:credit_ablation} shows consistent improvements as the planner components are introduced. Two-stage planning and matchability feedback each improve performance, while conditional credit assignment yields the largest gain, supporting separate credit for decomposition and collaboration decisions. \textsc{Decomposition + Single} still achieves an overall average of 50.4, showing that the learned decomposition alone is already strong, while learned Stage-II collaboration provides additional gains. 
Finally, the cost-aware objective reduces worker calls from 7.78 to 5.35 with only a modest decrease in average performance from 51.9 to 51.5, substantially improving the quality--cost trade-off.

\begin{table}[htbp]
\centering
\caption{
Ablation of the planner. 
}
\small
\setlength{\tabcolsep}{4pt}
\begin{tabular}{lcccc}
\toprule
Variant
& ID Avg. & OOD Avg. & Avg. & Calls $\downarrow$ \\
\midrule
Joint Planning
& 43.2 & 50.2 & 48.5 & 6.29 \\
Two-Stage
& 43.8 & 50.9 & 49.2 & 7.33 \\
\quad + Matchability
& 44.3 & 51.4 & 49.6 & 7.52 \\
\quad + Conditional Credit
& 47.7 & 53.4 & 51.9 & 7.78 \\
Decomposition + Single
& 46.1 & 51.8 & 50.4 & 1.68 \\
\method
& 47.4 & 52.9 & 51.5 & 5.35 \\
\bottomrule
\end{tabular}
\label{tab:credit_ablation}
\end{table}

%

\subsection{Analysis of the Matcher}
\label{sec:matcher_analysis}


\paragraph{Offline worker matching.}


We evaluate the static matcher independently of the learned planner. For in-distribution evaluation, examples from each probe source are randomly split 80/20 for probe construction and held-out evaluation. We additionally evaluate leave-one-source-out (LOSO) generalization on five representative sources by excluding the target source from the probe set. We compare against random routing, the strongest single worker, majority voting over all six workers, probe matching without density correction, and \textsc{Oracle Single}, which selects a successful worker in hindsight as an upper bound.
Table~\ref{tab:offline_matcher} shows that the full probe matcher achieves the strongest non-oracle performance in both settings. Its advantage narrows under LOSO-OOD evaluation, where no probe examples from the target source are available, but it still outperforms the strongest single-worker reference. Density correction provides consistent gains in both settings, supporting its role in reducing the influence of locally redundant probes. 
Finally, the sizable gap to \textsc{Oracle Single} indicates substantial remaining headroom in worker selection.

\begin{table}[htbp]
\centering
\caption{
Offline evaluation of worker matching.
}
\small
\setlength{\tabcolsep}{5pt}
\begin{tabular}{lccc}
\toprule
Method
& ID Avg. $\uparrow$
& LOSO-OOD Avg. $\uparrow$
& Calls $\downarrow$ \\
\midrule
Random
& 59.44 & 69.73 & 1 \\
\textsc{Best Single Worker}
& 68.33 & 75.26 & 1 \\
Majority vote
& 67.92 & 74.59 & 6 \\
Probe Matcher w/o Density
& 69.70 & 75.36 & 1 \\
Probe Matcher
& 70.24 & 75.45 & 1 \\
\textsc{Oracle Single}
& 82.68 & 87.17 & 1 \\
\bottomrule
\end{tabular}
\label{tab:offline_matcher}
\end{table}

\begin{table}[!ht]
\centering
\caption{
Online adaptation with increasing deployment feedback.
}
\small
\setlength{\tabcolsep}{7pt}
\begin{tabular}{lcccccc}
\toprule
Adaptation Examples
& 0
& 3K
& 6K
& 9K
& 12K
& 15K \\
\midrule
Overall Avg.
& 51.3 & 51.7 & 52.1 & 52.1 & 52.6 & 52.9 \\
\bottomrule
\end{tabular}
\label{tab:online_adaptation}
\end{table}

\paragraph{Online adaptation.}

We finally evaluate whether contextual bandit adaptation improves the matcher's worker-selection ability during deployment. We split the evaluation data into an adaptation stream and a disjoint held-out evaluation set, with the split detailed in Appendix~\ref{app:online_split}. The planner and the probe-based component of the matcher remain fixed throughout the experiment. Adaptation examples are processed sequentially, and the shared Thompson-sampling posterior is updated after each execution using the observed terminal outcome. After every 3,000 adaptation examples, we evaluate the current matcher on the fixed held-out set. Performance is computed separately on each benchmark and macro-averaged across the eight benchmarks. The zero-example checkpoint corresponds to the static probe matcher without deployment feedback.
Table~\ref{tab:online_adaptation} shows that performance generally improves as deployment feedback accumulates, increasing from 51.3 without online adaptation to 52.9 after 15K adaptation examples. This result demonstrates the effectiveness of our contextual bandit in progressively improving worker selection during deployment.


\section{Conclusion}

We introduced \method, a multi-agent orchestration framework that decouples worker-agnostic planning from concrete worker selection and uses two-stage planning for separate credit assignment to decomposition and collaboration decisions. Across diverse tasks, \method outperforms fixed and learned MAS baselines and remains effective under worker-pool changes without retraining. 
%

\subsection*{AI use statement}

Generative AI tools were only used to assist with language editing and polishing of the manuscript, including improving clarity, readability, grammar, and phrasing. All AI-assisted edits were reviewed by the authors. The authors take full responsibility for the final content of this work.






\bibliography{iclr2027_conference}
\bibliographystyle{iclr2027_conference}

\appendix


\section{Matcher Details}

\subsection{Online Matcher Adaptation}
\label{app:online_matcher}

We implement deployment-time adaptation using linear contextual Thompson
sampling with aggregate execution feedback. Online adaptation changes only
concrete worker selection; the planner-facing matchability features remain
probe-derived.

Let $z(\cdot)\in\mathbb{R}^{d}$ denote the frozen embedding produced by the
text encoder for either a subtask or a probe. Using the probe-density statistic
$\bar q_j$ defined in Section~\ref{sec:matcher}, we define a density-correction
weight
\[
    \rho_j
    =
    (\bar q_j+\epsilon)^{-\nu},
    \qquad
    Z
    =
    \sum_{j=1}^{N}\rho_j.
\]
For each worker $w$, its probe-derived capability representation is
\[
    g(w)
    =
    \frac{1}{Z}
    \sum_{j=1}^{N}
    \rho_j\,o_{w,j}\,z(p_j),
\]
where $p_j$ denotes probe $j$. The density correction reduces the influence of
semantically redundant probe regions, while the worker-independent
normalization $Z$ preserves differences in overall probe performance across
workers.

For a subtask--worker pair, we define the dimension-aligned interaction
\[
    \phi(s,w)
    =
    z(s)\odot g(w).
\]
Rather than maintaining a separate online parameter for every worker, we use a
single shared parameter $\eta\in\mathbb{R}^{d}$. Given a posterior sample
$\tilde\eta$, the deployment-adapted suitability is
\[
    \tilde u_w(s)
    =
    \hat u_w(s)
    +
    \tilde\eta^\top \phi(s,w).
\]
The adapted score replaces $\hat u_w(s)$ in the operation-specific
worker-selection rules, while probe-derived complementarity statistics remain unchanged.

For execution $t$, let $B_{t,s}$ denote the workers assigned to subtask $s$ and
\[
    A_t
    =
    \sum_s |B_{t,s}|.
\]
We aggregate the task--worker interactions and static suitability estimates of
the selected assignment as
\[
    x_t
    =
    \frac{1}{A_t}
    \sum_s
    \sum_{w\in B_{t,s}}
    \phi(s,w),
    \qquad
    \bar u_t
    =
    \frac{1}{A_t}
    \sum_s
    \sum_{w\in B_{t,s}}
    \hat u_w(s).
\]
We model the terminal utility as
\[
    r_t
    =
    \bar u_t
    +
    x_t^\top\eta
    +
    \epsilon_t,
    \qquad
    \epsilon_t\sim\mathcal N(0,\sigma^2).
\]

We place a zero-mean Gaussian prior
\[
    \eta
    \sim
    \mathcal N(0,\Lambda_0^{-1}).
\]
Treating $\bar u_t$ as a known probe-based offset, we define
\[
    y_t
    =
    r_t-\bar u_t
\]
and perform the standard Bayesian linear-regression update
\[
    \Lambda_t
    =
    \Lambda_{t-1}
    +
    \sigma^{-2}x_tx_t^\top,
    \qquad
    b_t
    =
    b_{t-1}
    +
    \sigma^{-2}x_ty_t,
\]
with posterior
\[
    p(\eta\mid\mathcal H_t)
    =
    \mathcal N
    \left(
        \Lambda_t^{-1}b_t,
        \Lambda_t^{-1}
    \right).
\]
Before each execution, Thompson sampling draws
$\tilde\eta\sim p(\eta\mid\mathcal H_t)$ for worker selection.

When a new worker joins, its probe signature immediately provides both
$\hat u_w(s)$ and $g(w)$. No worker-specific online parameter is introduced,
and the shared posterior over $\eta$ remains unchanged.
\section{Experimental Details and Additional Results}

\subsection{Dataset Statistics.}
\label{app:dataset}

Table~\ref{tab:data_statistics} summarizes the datasets and splits used throughout our experiments. Note that the probe set contains 55,676 examples. Profiling a new worker therefore requires roughly 56K probe inferences, which is typically inexpensive in practice and is incurred only when the worker is introduced. Since worker updates are far less frequent than deployment queries, the profiling cost is amortized over the worker's subsequent lifetime and adds little recurring overhead.

\begin{table*}[htbp]
\centering
\caption{
Data sources used for probe construction, planner post-training, and evaluation.
}
\label{tab:data_statistics}

\setlength{\tabcolsep}{5pt}
\begin{tabular}{lllcc}
\toprule
\textbf{Usage} &
\textbf{Dataset} &
\textbf{Primary task type} &
\textbf{\# Examples} &
\textbf{Sampling} \\
\midrule

\multirow{8}{*}{Probe set}
& APPS
& Code generation
& 4,962
& Full \\

& BFCL
& Tool / function use
& 3,151
& Full \\

& Big-Math
& Mathematical reasoning
& 6,000
& Sampled \\

& FRAMES
& Knowledge-intensive reasoning
& 824
& Full \\

& MBPP+
& Code generation
& 378
& Full \\

& SuperGPQA
& General knowledge / reasoning
& 26,529
& Full \\

& MMLU-Pro
& General knowledge / reasoning
& 12,032
& Full \\

& ACPBench
& Planning / agentic reasoning
& 1,800
& Full \\
\midrule

\multirow{8}{*}{RL training}
& DAPO-Math-17K
& Mathematical reasoning
& 1,000
& Sampled \\

& MuSiQue
& Multi-hop reasoning
& 675
& Sampled \\

& BigCodeBench
& Code generation
& 415
& Sampled \\

& Logistics
& Planning
& 400
& Sampled \\

& CodeContests
& Competitive programming
& 335
& Sampled \\

& TACO
& Code generation
& 335
& Sampled \\

& BrowseComp-Plus
& Retrieval / web reasoning
& 400
& Sampled \\

& TravelPlanner
& Planning
& 225
& Sampled \\
\midrule

\multirow{8}{*}{Evaluation}
& MuSiQue
& Multi-hop reasoning
& 2,000
& Sampled \\

& TravelPlanner
& Planning
& 1,000
& Full \\

& Omni-MATH-2
& Mathematical reasoning
& 4,428
& Full \\

& LiveCodeBench v6
& Competitive programming
& 611
& Full \\

& BBEH
& Broad reasoning
& 4,520
& Full \\

& NaturalPlan
& Constraint-based scheduling
& 3,600
& Full \\

& PlanBench-mystery
& Obfuscated symbolic planning
& 603
& Full \\

& GAIA
& General assistant / tool use
& 165
& Full \\
\bottomrule

\end{tabular}
\end{table*}

\subsection{Online Adaptation Split}
\label{app:online_split}

For the online adaptation experiment, we randomly hold out examples from each evaluation benchmark to construct a fixed evaluation set. Because GAIA contains only 165 examples, we reserve the full benchmark for evaluation. The remaining examples form the adaptation stream used to update the contextual-bandit posterior.

\begin{table}[htbp]
\centering
\caption{
Dataset split for the online adaptation experiment.
The held-out examples are used only for evaluation and never for posterior updates.
}
\small
\setlength{\tabcolsep}{8pt}
\begin{tabular}{lrrr}
\toprule
Benchmark
& Total
& Adaptation
& Held-out \\
\midrule
MuSiQue
& 2,000 & 1,750 & 250 \\
TravelPlanner
& 1,000 & 800 & 200 \\
Omni-MATH-2
& 4,428 & 4,128 & 300 \\
LiveCodeBench v6
& 611 & 411 & 200 \\
BBEH
& 4,520 & 4,220 & 300 \\
NaturalPlan
& 3,600 & 3,300 & 300 \\
PlanBench-mystery
& 603 & 391 & 212 \\
GAIA
& 165 & 0 & 165 \\
\midrule
Total
& 16,927 & 15,000 & 1,927 \\
\bottomrule
\end{tabular}
\label{tab:online_split}
\end{table}

\subsection{Implementation Details}
\label{app:details}

\paragraph{Planner training.}
We optimize the planner with a GRPO-style policy-gradient objective, applying $A^{\mathrm{dec}}$ to the Stage-I generation and $A^{\mathrm{op}}$ to the corresponding Stage-II generation, with the two losses weighted equally.
We perform full-parameter RL post-training of the planner with AdamW using a learning rate of $1\times10^{-6}$ under a cosine schedule with warmup ratio $0.03$, a global batch size of $16$ queries, and 400 optimization steps. For each training query, we sample $J=8$ Stage-I decompositions and $k=8$ Stage-II collaboration plans per decomposition, both at temperature $1.0$. Group-relative advantages are mean-centered within each group and are not divided by the group standard deviation. We allow at most $8$ subtasks per decomposition and cap the number of workers assigned to any single subtask at $K=6$. The Stage-II cost coefficient is set to $\beta=0.01$.

For planner post-training, terminal quality $Q(x,y)$ is computed by the task-specific verifier of each training source. For every source except TravelPlanner, the verifier is binary, with $Q\in\{0,1\}$ and $Q=1$ iff the generated solution passes. TravelPlanner instead uses a denser gated reward derived from the benchmark's own checkers: an invalid or undelivered plan receives zero reward, while a valid plan is scored as the equally weighted sum of five components---the fractions of commonsense and hard constraints satisfied, the corresponding two indicators of whether all such constraints are satisfied, and the benchmark's final all-or-nothing success indicator. We use this denser reward only for training because TravelPlanner's binary pass rate on our worker pool is near zero, causing rollout groups with uniformly zero rewards to provide no learning signal. Evaluation follows the standard task-specific grader of each benchmark, and all reported TravelPlanner results use the binary pass rate.

\paragraph{Static matcher.}
For probe-based matching, we use $h_{\mathrm{rel}}=0.05$, $h_{\mathrm{den}}=0.02$, $\nu=1.0$, and $\epsilon=0.01$. 
For MMR-based worker selection, we set the trade-off coefficient to
$\lambda_o=0.5$ for all multi-worker operations.

\paragraph{Online bandit.}
For online adaptation, we use the contextual Thompson-sampling formulation in Appendix~\ref{app:online_matcher}. We use the full 4096-dimensional \texttt{Qwen3-Embedding-8B} representation for both subtask embeddings $z(s)$ and probe embeddings $z(p_j)$. The shared online parameter $\eta$ uses an isotropic Gaussian prior with precision $1.0$, and the observation-noise variance is $\sigma^2=0.01$. 
Unless otherwise specified, all reported results use the static probe-based matcher without online updates; online adaptation is enabled only in the dedicated analysis in Table~\ref{tab:online_adaptation}.

\paragraph{Unseen worker pool.}
For the worker-pool generalization experiment in
Section~\ref{sec:pool_generalization}, we replace the original six-worker pool
with an entirely unseen pool of four models:
\texttt{gemma-4-26B-A4B-it}, \texttt{Qwen3-14B},
\texttt{granite-4.2-8b}, and
\texttt{Ministral-3-8B-Instruct-2512}. None of these workers is used during
planner training. For \method, each new worker is incorporated only by
evaluating it on the fixed probe set; no planner or matcher retraining is
performed. For \textsc{Conductor}, the four new workers are assigned the
available worker indices and the orchestrator is informed that the current pool
contains four workers, without further training. \textsc{Best Single Agent}
uses \texttt{gemma-4-26B-A4B-it}, the strongest worker in the new pool.
\textsc{Independent} executes all four workers independently and uses the
strongest worker to aggregate their responses into the final answer. Because
the current matcher forms teams from distinct workers, the default maximum team
size $K=6$ is clipped to the current pool size when fewer workers are available;
the four-worker unseen pool therefore allows at most four workers per subtask.
We omit MAS-Orchestra from the unseen-pool evaluation because it assumes a single underlying worker rather than selecting among a heterogeneous worker pool. Replacing this worker with one model from the new pool would test model substitution rather than adaptation to the pool, while introducing multi-worker selection would alter the original method.

\paragraph{Baseline adaptation.}
For \textsc{Conductor}~\citep{nielsen2026learning}, we replace its original worker pool with our six-worker pool and retrain the orchestrator following its original procedure. Although \textsc{Conductor} improves robustness to varying worker availability by training on randomized worker subsets, these subsets are drawn from the same fixed set of workers seen during training; its reported dynamic-pool evaluation therefore does not test transfer to entirely unseen workers. For \textsc{MAS-Orchestra}~\citep{ke2026mas}, which uses a single underlying worker model, we replace that worker with \texttt{gemma-4-12B-it}, the strongest worker in our pool, while otherwise retaining its original orchestration mechanism.
To approximately match the training rollout budget of \method
($400\times16\times8\times8\approx410$K executed orchestrations), we train \textsc{Conductor} for 1,600 optimization steps and \textsc{MAS-Orchestra} for 200 steps, accounting for their different numbers of rollouts per step.

All experiments are conducted on a single node with 4$\times$ NVIDIA H200 GPUs.
Code is available at: https://anonymous.4open.science/r/DeOrch-code-F8D3.

\subsection{Offline Matcher Evaluation Protocol}
\label{app:offline-matcher}

For the in-distribution evaluation of the static matcher, examples from each probe source are randomly split 80/20, with 80\% used to construct the probe set and the remaining 20\% held out for evaluation. For leave-one-source-out (LOSO) evaluation, we consider five representative sources---ACPBench, MMLU-Pro, MBPP+, Big-Math, and BFCL---and exclude the corresponding source entirely from the probe set before evaluating on its held-out examples.

We compare against random routing, the strongest individual worker in the pool (\texttt{gemma-4-12B-it}), majority voting over all six workers, and probe matching without density correction. We additionally report \textsc{Oracle Single}, which selects a successful worker for each example in hindsight and serves only as an upper bound.

\subsection{Complementarity-aware team selection.}
The previous experiment evaluates single-worker selection, while multi-worker operations additionally require constructing worker teams. To test whether probe-derived complementarity improves team selection, we construct a \textsc{Top-$n$ Suitability} variant of \method. Using the same planner outputs and team sizes as the full method, this variant selects the top-$n$ workers solely according to their suitability scores $\hat u_w(s)$, without using complementarity information.
Table~\ref{tab:binding_ablation} shows that complementarity-aware selection consistently improves over selecting workers solely by individual suitability. Since both variants execute the same plans with the same team sizes, the improvement cannot be attributed to additional worker calls, but instead reflects the benefit of accounting for worker complementarity when constructing multi-worker teams.

\begin{table}[htbp]
\centering
\caption{
Inference-time ablation of multi-worker selection.
}
\small
\setlength{\tabcolsep}{6pt}
\begin{tabular}{lccc}
\toprule
Selection Rule
& ID Avg. $\uparrow$
& OOD Avg. $\uparrow$
& Overall Avg. $\uparrow$ \\
\midrule
\textsc{Top-$n$ Suitability}
& 45.9 & 52.3 & 50.7 \\
Complementarity-Aware selection
& 47.4 & 52.9 & 51.5 \\
\bottomrule
\end{tabular}
\label{tab:binding_ablation}
\end{table}

\subsection{Inference Cost Analysis}
\label{app:cost_analysis}

We further evaluate inference cost by counting total prompt and completion tokens across all autoregressive model calls, including planning, worker execution, and collaboration. Token usage is averaged within each benchmark and macro-averaged across the eight benchmarks. We additionally include \textsc{Best Single + Refinement} to control for increased inference compute: starting from \texttt{gemma-4-12B-it}, it performs three sequential refinement rounds without ground-truth feedback, yielding four worker calls and a token budget comparable to \method.

Table~\ref{tab:inference_cost} shows that refinement improves \textsc{Best Single Agent} from 39.2 to 42.4, but remains well below \method's 51.5 at a similar token budget (14.1K vs.\ 14.6K). \method also outperforms \textsc{Independent}, \textsc{Conductor}, and \textsc{MAS-Orchestra} while using fewer tokens, indicating that its gains are not explained by increased inference compute alone.

\begin{table}[htbp]
\centering
\caption{
Inference cost across the eight evaluation benchmarks.
}
\small
\setlength{\tabcolsep}{6pt}
\begin{tabular}{lccc}
\toprule
Method
& Avg. $\uparrow$
& Calls $\downarrow$
& Tokens / Query $\downarrow$ \\
\midrule

\textsc{Best Single Agent}
& 39.2 & 1.00 & 1859 \\

\textsc{Best Single + Refinement}
& 42.4 & 4.00 & 14106 \\

\textsc{Independent}
& 46.0 & 7.00 & 25859 \\

\textsc{Frontier-Planner}
& 36.6 & 4.67 & 11479  \\

\textsc{Untrained Planner}
& 38.3 & 4.92 & 12867 \\

\textsc{Conductor}
& 44.0 & 6.16 & 22458 \\

\textsc{MAS-Orchestra}
& 49.1 & 5.63 & 17250 \\

\method
& 51.5 & 5.35 & 14577 \\

\bottomrule
\end{tabular}
\label{tab:inference_cost}
\end{table}

\subsection{Comparison with Counterfactual Credit Assignment}

LEMON addresses orchestration credit assignment through counterfactual edits to its generated orchestration. We do not include it as a direct baseline in Table~1 because LEMON assigns each role to one of three predefined capacity levels (\textit{small}, \textit{medium}, or \textit{large}), whereas our setting requires selecting freely among six heterogeneous workers. Adapting LEMON to this setting would require redefining both its worker-selection representation and the corresponding counterfactual edits, making the comparison no longer a faithful reproduction of the original method.

We instead construct a controlled LEMON-inspired counterfactual-credit baseline in our orchestration space. 
This variant replaces our two-stage planner and conditional-credit objective with single-pass orchestration trained using LEMON-style global and localized counterfactual objectives, while keeping the worker pool, matcher, execution protocol, and training budget unchanged. Specifically, following LEMON's counterfactual training procedure, we train the single-pass planner with global GRPO augmented by one localized counterfactual edit per rollout; we adapt its mutations to our orchestration space by editing either a decomposition dependency or a collaboration operation, re-execute the modified plan, and apply the resulting reward contrast only to the edited token span. This directly compares two approaches to credit assignment across decomposition and operation selection: counterfactual credit assignment versus our two-stage conditional formulation. As shown in Table~\ref{tab:counterfactual}, \method improves the overall average from 49.2 to 51.5 with comparable worker calls, indicating that our two-stage conditional formulation provides more effective credit assignment in this setting.

\begin{table}[htbp]
\centering
\caption{Comparison of credit-assignment formulations.}
\small
\setlength{\tabcolsep}{4pt}
\begin{tabular}{lcccc}
\toprule
Variant
& ID Avg. & OOD Avg. & Avg. & Calls $\downarrow$ \\
\midrule
Counterfactual Credit
& 43.1 & 51.2 & 49.2 & 5.17 \\
Two-Stage Conditional Credit (\method)
& 47.4 & 52.9 & 51.5 & 5.35 \\
\bottomrule
\end{tabular}
\label{tab:counterfactual}
\end{table}


\subsection{Discussion and Future Work}
\label{app:discussion}

\paragraph{Computational overhead.}
Compared with single-pass planning, our two-stage planner introduces one additional planner generation at inference time. This overhead is modest relative to the subsequent execution of multiple worker agents; indeed, our cost analysis shows that \method still uses fewer inference tokens than competing learned MAS orchestrators. Training incurs a larger overhead because the nested conditional rollout procedure samples multiple Stage-II plans for each sampled decomposition. However, this additional training cost is incurred only during post-training: once trained, the same planner can be reused when the worker pool changes without further retraining. Improving the efficiency of this nested training procedure is an interesting direction for future work.

\paragraph{Extending the collaboration space.}
Our current multi-worker operations construct teams from distinct workers in the pool and therefore do not support homogeneous collaboration through repeated execution of the same worker. In practice, an \textsc{Ensemble}-3 operation could either invoke three distinct workers or run three independent instances of the same strong worker, and which choice is preferable may depend on the task. A natural solution is to introduce both homogeneous and heterogeneous variants into the operation vocabulary, such as homogeneous-\textsc{Ensemble}-3 and heterogeneous-\textsc{Ensemble}-3, allowing the planner to learn which form of collaboration is more appropriate for each subtask. We also considered richer operations that are not explored in this work due to computational constraints. A recursive operation could further decompose a difficult subtask during execution, while an escalation operation could invoke an external worker outside the current pool when the available workers are insufficient. Incorporating such operations would extend the framework toward more adaptive and open-ended orchestration.

\paragraph{Scaling to larger worker pools.}
Our experiments consider heterogeneous pools of up to six workers. Although the planner interface is independent of worker identities and pool size, the behavior of the matcher at substantially larger scales remains to be evaluated. Evaluating and improving \method under substantially larger and more dynamic worker pools is therefore an important direction for future work.

\end{document}

%% file: math_commands.tex
\usepackage{amsmath,amsfonts,bm}

\def\eqref#1{equation~\ref{#1}}

\def\1{\bm{1}}

\DeclareMathAlphabet{\mathsfit}{\encodingdefault}{\sfdefault}{m}{sl}
\SetMathAlphabet{\mathsfit}{bold}{\encodingdefault}{\sfdefault}{bx}{n}

